# KNET: Integrating Hypermedia and Bayesian Modeling

R. Martin Chavez and Gregory F. Cooper
Medical Computer Science Group
Stanford University
Stanford, California, U.S.A. 94305

## Abstract

KNET is a general-purpose shell for constructing expert systems based on belief networks and decision networks. Such networks serve as graphical representations for decision models, in which the knowledge engineer must define clearly the alternatives, states, preferences, and relationships that constitute a decision basis. KNET contains a knowledge-engineering core written in Object Pascal and an interface that integrates Hyper-Card, a hypertext authoring tool for the Apple Macintosh computer, into an expert-system architecture. Hypertext and hypermedia have become increasingly sophisticated in their storage, management, and retrieval of information. In broad terms, hypermedia deliver heterogeneous bits of information in dynamic, extensively cross-referenced packages. The resulting KNET system features a coherent probabilistic scheme for managing uncertainty, an object-oriented graphics editor for drawing and manipulating decision networks, and HyperCard's potential for quickly constructing flexible and friendly user interfaces. We envision KNET as a useful prototyping tool for ongoing research on a variety of Bayesian reasoning problems, including tractable representation, inference, and explanation.

## 1 Motivation

### 1.1 User Interfaces

Rowley *et al.* observe that "advances in computer science are often consolidated as programming systems that raise the abstraction level and the vocabulary for expressing solutions to new problems. We have seen little permanent consolidation of this form in AI" [Rowley, 1987]. The authors note that some artificial intelligence (AI) systems place too many restrictions on the admissible paradigms; almost all insulate the AI kernel from the surrounding programming environment; few support the inclusion of facilities that were not coded within the original framework. Our experience has illustrated shortcomings in many of the available knowledge-engineering products, including EMYCIN [Buchanan and Shortliffe, 1984], KEE [Intellicorp, Inc., 1986], S-1 [Teknowledge, Inc., 1984], and Personal Consultant Plus. Consultations with knowledge bases developed in EMYCIN and Personal Consultant Plus must conform to a rigidly specified, linear sequence of questions and answers. The certainty factor model of EMYCIN, Personal Consultant Plus, and S-1 implicitly assumes that rules must form tree-structured chains of inference. KEE provides no facilities for the management of uncertainty.

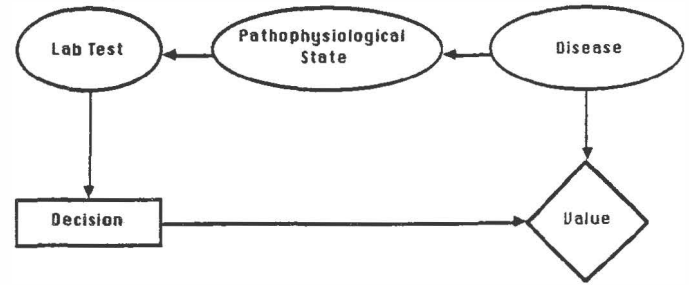

Figure 1: THIS DECISION NETWORK CAPTURES GENERIC KNOWLEDGE ABOUT A MEDICAL DECISION BASED ON THE OUTCOME OF A LABORATORY TEST. AN ALGORITHM FOR DECISION NETWORKS WOULD ANSWER THE QUESTION: WHICH DECISION YIELDS THE HIGHEST EXPECTED VALUE FOR THIS PATIENT?

KNET combines normative probabilistic modeling techniques with a front end that offers the flexibility and expressiveness of a hypertext system. Perhaps more important, KNET separates the design of a tailored, domain-specific user interface from all other aspects of the system. In addition, KNET strictly adheres to the Macintosh human-interface guidelines. Buttons, icons, scrolling text fields, color illustrations, menus, and mouse-sensitive screen objects can be applied to build knowledge-acquisition interfaces that facilitate the construction and validation of new knowledge bases. HyperCard, an authoring tool for hypermedia, facilitates the design of intuitive user interfaces. A custom-designed interprocess communications channel transfers information from the object-oriented KNET core (written in Object Pascal) to HyperCard and back.

The KNET environment runs on low-cost, general-purpose hardware. Using HyperTalk [Goodman, 1987], Hyper-Card's object-oriented authoring language, knowledge engineers as well as relatively naive users can incorporate sound, synthesized speech, videodisc images, and animation into their applications. Our experience has indicated that we can prototype, debug, and refine substantially different user interfaces to a Bayesian model in a single session. Other workers in the field [Buchanan, 1988] have observed that powerful and understandable user interfaces can absorb as much as 80% of a project's design and implementation cycle. We believe that KNET enables the development of an appropriate expert-system interface interface that takes much less of the development cycle than heretofore possible.

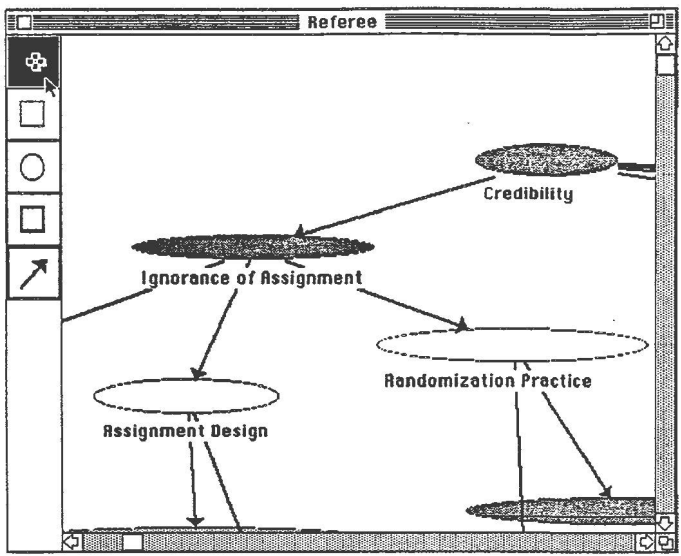

Figure 2: A SMALL PORTION OF THE REFEREE BELIEF NETWORK, AS DISPLAYED BY KNET. REFEREE ENCODES KNOWLEDGE ABOUT THE INTERPRETATION OF RANDOMIZED CLINICAL TRIALS.

## 1.2 Belief networks and decision networks

Belief networks and decision networks have been used as tools for constructing coherent probabilistic representations of uncertain expert opinion [Henrion, 1987]. A decision network is a type of influence diagram used for decision making [Howard and Matheson, 1984]; we use the term specifically to designate a well-formed decision influence diagram [Holtzman, 1988]. For more than a decade now, decision analysts have used decision networks to construct formal descriptions of decision problems and to capture knowledge in a representation that people with varying degrees of technical proficiency can understand. Belief networks are specialized decision networks that lack contain only chance nodes. Belief networks are particularly useful for diagnostic applications. Pearl [Pearl, 1986] has proposed an elegant distributed algorithm for belief maintenance and updating in such networks.

Decision networks represent the alternatives, states, preferences, and relationships that constitute a decision model [Howard and Matheson, 1984]. We define decision networks as directed acyclic graphs with nodes that represent propositions or quantities of interest and arcs that summarize the interactions between those nodes. Decision networks provide a mapping between the expert's knowledge and the internal computational formalism. Their intuitive pictorial structure facilitates knowledge acquisition and communication [Horvitz et al., 1988].

The simple network of Figure 1 encodes a generic problem in medical decision-making. Rectangular decision nodes represent actions under direct control of the decision maker. In this case, the physician and the patient must decide whether to undertake a course of treatment. Arcs that enter a decision node represent the information available at the time of action. Circular chance nodes (for instance, the

PATHOPHYSIOLOGICAL STATE of the patient, the presence or absence of an underlying DISEASE, and the result of a LABORATORY TEST) represent uncertain states of the world. Diamond-shaped value nodes summarize the preferences of the decision maker. In the example of Figure 1, the net value might depend on life expectancy, quality of life, and costs associated with diagnostic and treatment intervention. The decision maker can use the network to determine a course of action that maximizes expected value.

After drawing a decision network, the knowledge engineer must quantitate the influences of parent nodes on their children. Chance nodes without predecessors require prior-probability distributions. Chance nodes with predecessors require probability distributions conditioned on their parents. Next, the knowledge engineer must encode the decision maker's attitudes toward risk according to the axioms of utility theory. Value nodes require the specification of a function over all its parent decision and chance nodes. An inference algorithm for belief networks will calculate posterior odds, based on all the available evidence, for each chance node of interest. An inference algorithm for decision networks will determine the decisions with the highest expected value.

Shachter has designed DAVID, a decision network processing system that runs on the Macintosh and provides operations for expected-value decision making and sensitivity analysis [Shachter, 1986]. He observes that "the criticisms of probabilistic models of uncertainty are overcome by an intelligent graphical interface that explicitly incorporates conditional independence" [Shachter, 1986]. Shachter's encouraging results show that students have been able quickly to build and solve decision models with DAVID. Henrion [Henrion, 1987] has demonstrated the feasibility of constructing decision networks of moderate size (with about 30 nodes) for diagnosing and treating plant disease. MUNIN [Lauritzen and Spiegelhalter, 1988], an expert system based on belief networks for electromyography diagnosis, pursues a similar knowledge-engineering approach. Experience suggests, therefore, that belief networks and decision networks can serve as effective representations for communication between people and machines. The elicitation of those normative network models, moreover, entails a methodology that can assist in removing imprecise language, forcing clear explication of an expert's model, and clarifying the interrelationships of causal influences.

Typically, the development of practical skills, algorithms, and software lags behind the theoretical discovery of a new modeling paradigm. Shachter [Shachter, 1986] proposes an architecture that integrates decision networks and traditional expert systems. KNET is a step toward that goal. KNET provides graphical tools, HyperCard templates for defining user interfaces, a Bayesian decision-making kernel, and an open architecture that encompasses Object Pascal, HyperTalk, and in the near future, CommonLisp. The target user never needs to observe the details of uncertainty management in KNET, inasmuch as the hypermedia interface hides the irrelevant details.

The architectural features that distinguish KNET from other systems are as follows:

- KNET allows chance nodes and decision nodes to be instantiated to specific background values prior to per-

forming inference. By contrast, many other decision-analysis systems, such as DAVID, explore the results of many potential background states in order to gain insight into the model. Those systems focus on the problem of decision making; KNET supports diagnosis as well as decision making.

- KNET envisions a multiplicity of target audiences that require different interfaces. Hyptertext simplifies the authoring of new interfaces.

- KNET specifies an open architecture for probabilistic expert systems. A distributed server that runs on fast hardware could, for instance, perform updating of the decision model. A networked implementation will allow the Macintosh to offer maximum responsiveness, whereas superior number-crunching hardware will execute the inferencing algorithm.

# 2 Knowledge engineering of a Bayesian model

In its present form, KNET uses Pearl's distributed updating algorithm [Pearl, 1986] to maintain belief assignments in a belief network; the specific implementation is described in [Suermondt and Cooper, 1988]. The system accomodates decision nodes by using a technique that transforms any belief-network algorithm into a decision-network algorithm [Cooper, 1988].

Design of a belief network follows the canonical principles of decision analysis [Howard and Matheson, 1984]. First, the knowledge engineer must extract the relevant state variables and their admissible values. In certain domains, such as clinical epidemiology, where few terms possess a unique denotation, the engineer must elicit detailed descriptions that pass a clarity test. In other words, a clairvoyant with access to all the relevant information could unambiguously assign a value to each state variable without requiring further clarification.

Second, the knowledge engineer and expert must group the state variables into a directed acyclic graph by drawing arcs that represent influences. Arcs may, but do not necessarily, denote cause-and-effect relationships. The absence of an arc implies specific probabilistic statements of conditional independence. Pearl [Pearl, 1986] and Wellman [Wellman, 1986] describe the implications of graph connectivity in detail.

Anyone who has used standard Macintosh graphics software can create decision networks and belief networks in KNET. For example, after invoking a HyperCard stack entitled "Belief Networks," the engineer presses a button labeled "New Knowledge Base." Two empty windows appear: One displays a palette of graphical shapes, and the other contains a scrollable list of variable names. The engineer clicks the mouse on the palette to select ellipses for chance nodes, squares for decision nodes, diamonds for value nodes, and arrows for influence arcs. Using mouse gestures that the Macintosh design group has standardized, the knowledge engineer creates a color picture of the network (Figure 2). KNET supports all the sophisticated screen operations that Macintosh users expect, including dragging, coloring, shading, naming, and resizing of nodes. KNET's pop-up window scroller allows the engineer immedi-ately to situate herself within a potentially large knowledge base and to extract a view of any slice of the domain.

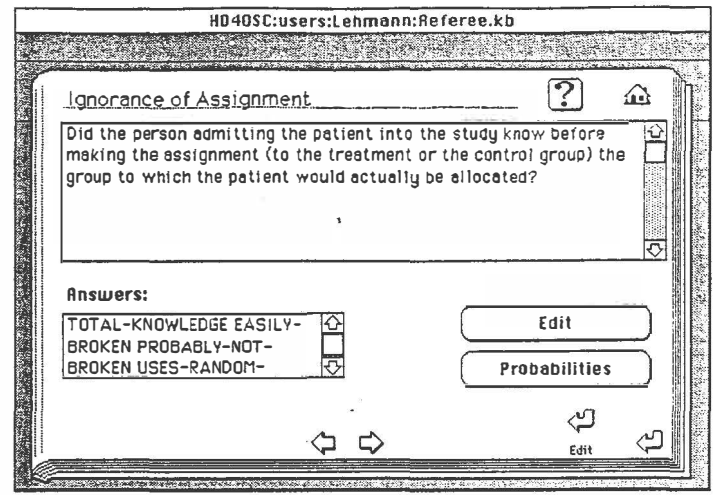

Figure 3: A HYPERCARD VIEW OF A CHANCE NODE, IGNORANCE OF ASSIGNMENT, FROM THE REFEREE BELIEF NETWORK. DIFFERENT KNOWLEDGE BASES MIGHT PRESENT DOMAIN-SPECIFIC HYPERTEXT RENDITIONS OF THEIR CONCEPTS.

After drawing a network and rearranging its nodes for maximum esthetic appeal, the engineer can double-click on a node and thereby open that node for further definition and inspection. A HyperCard corresponding to the node appears on the screen (Figure 3). At present, KNET provides only one template for the knowledge engineer's view of a node; we plan to offer several, inasmuch as the design of a new format with different text fields, buttons, labels, graphs, and illustrations takes only minutes, and presupposes no deep knowledge of HyperTalk programming.

The inferencing core for belief networks requires that the HyperCard interface specify mutually exclusive and collectively exhaustive values for each node, a prior belief assignment for those nodes that have no incoming arcs, and a conditional probability distribution over the values of the parent nodes. We have extended HyperCard to share hypertext fields with Pascal objects. The communication mechanism, although complex at the implementation level, presents aan interface that the HyperTalk script designer can invoke with concise messages. In effect, the Hyper-Card interface can create sophisticated views of the decision network's hidden structure. More important, the KNET programmer can tailor those views to the requirements of diverse knowledge engineers who are operating in different domains. Difficult graphical programming in Pascal or Lisp is never required; instead, the HyperTalk scripting language and drawing tools make rapid prototyping and refinement possible.

An example illustrates the key ideas. A prototypical HyperCard view (Figure 4) into a chance node of the belief network contains a scrolling text field labeled "QUESTION," a field labeled "LEGALVALUES," a "NAME" field at the top, a button labeled "PROBABILITIES," and a return arrow at the bottom. The return arrow contains a simple HyperTalk script with the command "activate KNET"; when the user clicks on that button, HyperCard returns control to the KNET knowledge-engineering code, and the belief network (Figure 3) becomes the foremost window on the screen. The "PROBABILITIES" button has a script that transfers control to another HyperCard (not shown), one that gathers numbers for the conditional probability distribution incident upon that node. The probability-gathering card has at its disposal all the computational, text-manipulation, and painting capabilities of HyperTalk. The card need only insert the new conditional probability distribution of a node into an invisible hypertext field; the interface shares that field transparently with Object Pascal.

HyperCard provides a number of useful templates. One of those prototype cards contains HyperTalk scripts that produce histograms and pie charts corresponding to arbitrary numerical data. With 5 minutes of effort, the first author was able to paste the chart-drawing card into his KB stack, to create a new button on the probability-gathering card, and to write a five-line script that switches between graphical and numeric views of the conditional probability distribution. Such a facility would require a significant amount of programming time; a requirement for user-specified dimensions, labels, and background illustrations could quickly turn into a tedious, lengthy task for the system's programmer. With KNET and HyperCard working together, however, the KNET designer need not anticipate every possible feature required by every conceivable target audience. KNET provides the templates and hooks; Hyper-Card users, be they domain experts, knowledge engineers, or application programmers, can do the rest.

## 3  Using the model

A fully specified belief network enumerates the possible values or discretized ranges for each state variable, the relevant conditional probability distributions at each local event group, and prior probabilities for the root nodes. A single line of HyperTalk code, "command consult," tells KNET to instantiate the belief network, to calculate evidential and diagnostic support for each node, to compute the current belief assignment, and to build a new consultation object. We have extended the EMYCIN metaphor of a decision session as a consultation with the expert's knowledge base; in KNET, however, the consultation can use hypertext to improve the flow of information between knowledge base and the user. We have enhanced Hyper-Talk with commands that extract the belief assignment for a given node and feed new observations into the belief network. The KNET architecture hides the irrelevant details of belief updating and propagation. From the user's point of view, observations automatically propagate through the network and make themselves apparent through graphical interactions with the HyperCard interface.

The current consultation format (Figure 4), which was designed and built in a few hours, allows the user to explore the belief network by pointing and clicking on the obvious icons. Transition from one card to the next happens instantaneously; on-screen buttons provide helpful information, illustrate the user's context within the model, and effect transitions among different levels of the model. One button provides easy access to a dictionary of terms; others break complicated queries into more easily managed parts. Labeled scales translate mouse gestures into belief updates. Buttons, when activated, reveal a node's current belief assignment in either numerical or graphical form. By pointing and double-clicking, the user can switch between color views of the belief network and hypertext presentations of the equivalent information; users may choose to have both displays available simultaneously. In short, the current version of KNET provides a consultation format that fully exploits progress in the design of human interfaces. In addition, KNET offers that generality and flexibility within a normative Bayesian framework.

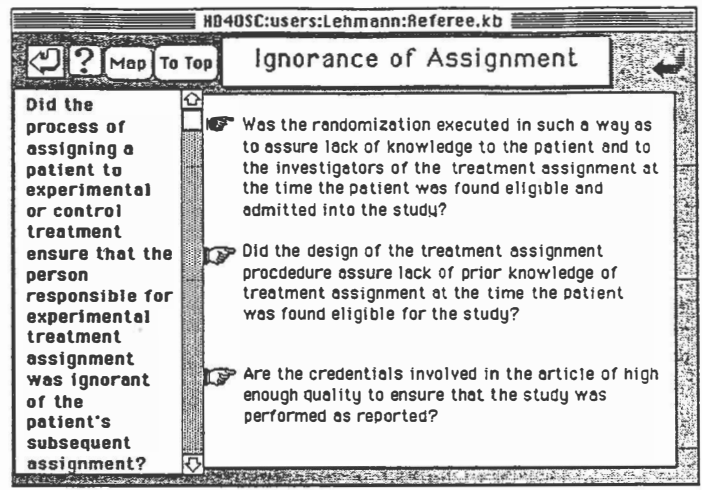

Figure 4: A SCREEN VIEW FROM A CONSULTATION WITH REFEREE SHOWS THE RELATIONSHIP BETWEEN THE BELIEF NETWORK AND THE HYPERCARD INTERFACE. BY CLICKING ON THE POINTING-HAND ICONS, THE USER CAN FOCUS THE SYSTEM'S ATTENTION FROM IGNORANCE OF ASSIGNMENT TO MORE SPECIFIC QUESTIONS ABOUT THE RANDOMIZATION OF THE CLINICAL STUDY.

## 4  Applications

REFEREE [Chavez and Lehmann, 1988] is an expert system that incorporates into a belief network a biostatistician's expert knowledge about the methodology of randomized clinical trials (RCTs). An original EMYCIN prototype revealed ambiguity in the goals of the project and in the precise definition of state variables (also known as "parameters" in EMYCIN). Perhaps more significant, the EMYCIN implementation assigned conflicting interpretations to certainty factors, which simultaneously served as measures of belief and as continuous measures of quality. In addition, EMYCIN's facilities for observing the interrelationships among rules and for guiding the user through a consultation with the expert's knowledge base were not adequate. Without advanced tools for clarifying questions and for providing the necessary contextual clues, experts and knowledge engineers became confused about the underlying structure and purpose of the system.

In parallel with the design and implementation of KNET on the Macintosh, we drew the REFEREE belief network and wrote scripts to transfer the old knowledge-base frames (from a prototype implementation on the TI Explorer) into HyperCard. To date, we have found that we can act on the REFEREE team's suggestions and can demonstrate an enhanced version of the HyperCard interface within hours to a few days. We have constructed a belief network that represents the REFEREE expert's subjective knowledge about the interpretation of randomized, controlled studies that measure the effect of a treatment intervention on mortality. We are now validating and adjusting the expert's numerical assessments.

## 5  Future work

KNET has begun to serve as our research group's vehicle for investigating the design of large knowledge-intensive systems with coherent schemes for managing uncertainty. We have planned and initiated the following activities:

- The general Bayesian inferencing problem is NP-hard [Cooper, 1987]. We are investigating randomized algorithms that may yield significant time reductions for networks with particular topologies. Inasmuch as KNET hides the details of belief propagation, we can develop, test, and incorporate new algorithms without altering existing knowledge bases.

  Research in progress [Chavez, 1988] suggests that Monte Carlo area-estimation strategies, combined with a rapidly convergent Markov chain that generates hypothetical scenarios, can (with high probability) estimate all posterior distributions to within a prespecified relative error.

- Belief networks can encode meta-knowledge about how to manage and focus a user's interaction with a Bayesian model [Horvitz, 1988]. KNET offers a general facility for designing and verifying belief networks; we can then package and reference those networks through extensions to HyperTalk. In the coming months, we will apply KNET to ongoing work on belief networks for control reasoning.

- Our expert has suggested the use of color as an explanation facility; nodes might be shaded or colored according to their influence on the goal node. Inasmuch as HyperCard can control the presentation of a decision network by setting red, green, blue tuples in hypertext fields, the basic facilities already exist. In consultation with our expert, we shall experiment with various color-coded semantics for explanation.

- KNET presently uses a primitive custom database to store large quantities of information, including numerical data, network topology, color coding, shading, and discretization. We are presently incorporating a relational-database management system (DBMS) into the KNET architecture [Barsalou and Wiederhold, 1988]. We shall access the DBMS from both Object Pascal and HyperCard. The DBMS will store risk-preference curves, probability distributions, pictorial data, incremental revisions of the knowledge base, and consultation histories.

- As a test of KNET's flexibility and general utility, we are converting the PATHFINDER knowledge base [Horvitz et al., 1984] to KNET format. PATHFINDER presently assumes conditional independence of evidence given diseases; the KNET implementation will allow us to relax that assumption. We shall measure the time required for the transition to KNET and the construction of an appropriate HyperCard consultation interface.

Only further experience will establish the efficacy of Bayesian methods in the design of large-scale expert systems. KNET offers the software tools needed to design, debug, and validate Bayesian models suitable for use by a large and diverse target audience.

## Acknowledgments

Harold Lehmann made essential contributions to the development of KNET. Diana Forsythe, David Heckerman, Eric Horvitz, Bruce Buchanan, and Lyn Dupre offered valuable suggestions. Bill Brown and Dan Feldman devoted their time and insight as domain experts.

This work was supported by the National Library of Medicine under Grant LM-04136, the National Institutes of Health under a Medical Scientist Training Program Grant GM-07365, and the National Science Foundation under Grant IRI-8703710. Computing facilities were provided in part by the SUMEX-AIM resource under NIH Grant RR-00785.